\documentclass[letterpaper, 10 pt, conference]{ieeeconf}  

\IEEEoverridecommandlockouts                              

\title{\LARGE \bf Learning Quickly to Plan Quickly Using Modular Meta-Learning}
\author{
  \textbf{Rohan Chitnis \hspace{16mm} Leslie Pack Kaelbling \hspace{8mm} Tom\'as Lozano-P\'erez}\\\\
  MIT Computer Science and Artificial Intelligence Laboratory\\
  \texttt{\{ronuchit, lpk, tlp\}@mit.edu}
  \thanks{\hspace*{-1em}Presented at the 2019 IEEE International Conference on Robotics and Automation (ICRA), Montreal, Canada.}
}

\usepackage{multicol}
\usepackage{color}
\usepackage{amssymb}
\usepackage{amsmath}
\usepackage{tikz}
\usetikzlibrary{arrows,decorations.markings}

\usepackage{booktabs}
\usepackage{graphicx}
\usepackage{todonotes}

\usepackage[font=small]{caption}
\usepackage{subcaption}

\usepackage[noline]{algorithm2e}
\makeatletter
\newcommand{\removelatexerror}{\let\@latex@error\@gobble}
\makeatother

\usepackage{indentfirst}
\usepackage{float}
\usepackage{mathtools}

\DeclareMathOperator*{\argmin}{argmin}

{\begin{list}{$\bullet$}{%
    \setlength{\topsep}{0in}
    \setlength{\partopsep}{0in}
    \setlength{\itemsep}{0in}
    \setlength{\parsep}{0in}
    \setlength{\leftmargin}{1.5em}
    \setlength{\rightmargin}{0in}
    \setlength{\itemindent}{-.1in}
}
}%
{\end{list}
}

\newcommand{\secref}[1]{Section~\ref{#1}}

\newcommand{\figref}[1]{Figure~\ref{#1}}

\newcommand{\tabref}[1]{Table~\ref{#1}}

\DeclarePairedDelimiterX{\infdivx}[2]{(}{)}{%
  #1\;\delimsize\|\;#2%
}

\newcommand{\I}{\mathcal{I}}
\newcommand{\G}{\mathcal{G}}

\newcommand{\St}{\mathcal{S}}
\newcommand{\A}{\mathcal{A}}

\newcommand{\pomdp}{{\sc pomdp}}
\newcommand{\maml}{{\sc maml}}

\newcommand{\Loss}{\mathcal{L}}

\newcommand{\tamp}{{\sc tamp}}
\newcommand{\spec}{specializer}
\newcommand{\Spec}{Specializer}
\newcommand{\bouncegrad}{{\sc BounceGrad}}
\newcommand{\moma}{{\sc moma}}
\newcommand{\data}{\mathcal{D}}
\newcommand{\subs}{{\sc ss}}
\newcommand{\ad}{{\sc ad}}
\newcommand{\procm}[1]{\text{\sc #1}}

\begin{document}
\bibliographystyle{IEEEtran}
\maketitle
\thispagestyle{empty}
\pagestyle{empty}
\setlength{\textfloatsep}{8pt}

\begin{abstract}
  Multi-object manipulation problems in continuous state and action
  spaces can be solved by planners that search over sampled values for
  the continuous parameters of operators. The efficiency of these
  planners depends critically on the effectiveness of the samplers
  used, but effective sampling in turn depends on details of the
  robot, environment, and task. Our strategy is to learn functions
  called \emph{specializers} that generate values for continuous
  operator parameters, given a state description and values for the
  discrete parameters. Rather than trying to learn a single
  specializer for each operator from large amounts of data on a single
  task, we take a {\em modular meta-learning} approach. We train on
  multiple tasks and learn a variety of specializers that, on a new
  task, can be quickly adapted using relatively little data -- thus,
  our system \emph{learns quickly to plan quickly} using these
  specializers. We validate our approach experimentally in simulated
  3D pick-and-place tasks with continuous state and action
  spaces. Visit \texttt{http://tinyurl.com/chitnis-icra-19} for a
  supplementary video.
\end{abstract}

\section{Introduction}
\label{sec:introduction}
Imagine a company that is developing software for robots to be
deployed in households or flexible manufacturing situations. Each of
these settings might be fairly different in terms of the types of
objects to be manipulated, the distribution over object arrangements, or
the typical goals. However, they all have the same basic underlying
kinematic and physical constraints, and could in principle be solved
by the same general-purpose task and motion planning (\tamp{})
system. Unfortunately, \tamp{} is highly computationally intractable
in the worst case, involving a combination of search in symbolic
space, search for motion plans, and search for good values for
continuous parameters such as object placements and robot
configurations that satisfy task requirements.

A robot faced with a distribution over concrete tasks can learn to
perform \tamp{} more efficiently by adapting its search strategy to
suit these tasks. It can learn a small set of typical grasps for the
objects it handles frequently, or good joint configurations for taking
objects out of a milling machine in its workspace. This distribution
cannot be anticipated by the company for each robot, so the best the
company can do is to ship robots that are equipped to learn very
quickly once they begin operating in their respective new workplaces.

The problem faced by this hypothetical company can be framed as one of
{\em meta-learning}: given a set of tasks drawn from some meta-level
task distribution, learn some structure or parameters that can be used
as a {\em prior} so that the robot, when faced with a new task drawn
from that same distribution, can learn very quickly to behave
effectively.

Concretely, in this work we focus on improving the interface between
symbolic aspects of task planning and continuous aspects of motion
planning.  At this interface, given a symbolic plan structure, it is
necessary to select values for continuous parameters that will make
lower-level motion planning queries feasible, or to determine that the
symbolic structure itself is infeasible.  Typical strategies are to
search over randomly sampled values for these parameters, or to use
hand-coded ``generators'' to produce them~\cite{tamphpn,tampinterface}.

Our strategy is to learn deterministic functions, which we call {\em
  \spec{}s}, that map a symbolic operator (such as {\em place(objA)})
and a detailed world state description (including object shapes,
sizes, poses, etc.) into continuous parameter values for the operator
(such as a grasp pose). Importantly, rather than focusing on learning
a single set of \spec{}s from a large amount of data at deployment
time, we will focus on meta-learning approaches that allow \spec{}s to
be quickly adapted online. We will use deep neural networks to
represent \spec{}s and backpropagation to train them.

\begin{figure}[t]
  \centering
    \noindent
    \includegraphics[width=0.35\columnwidth]{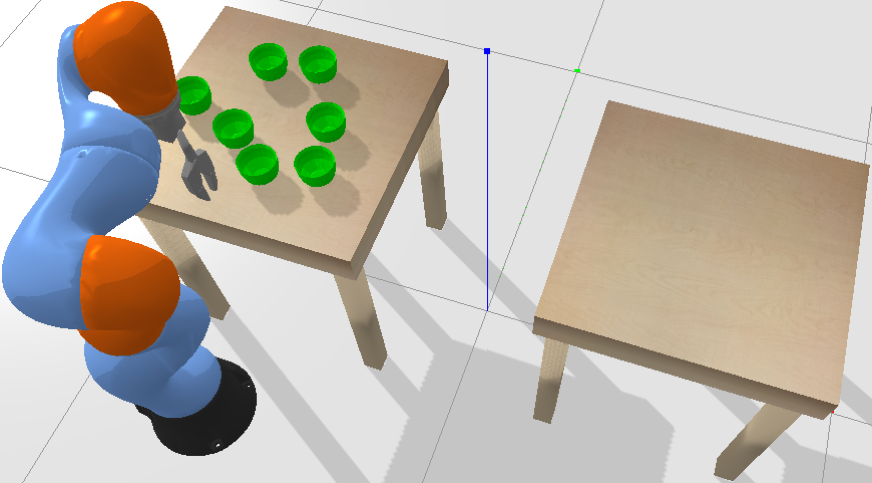}
    \includegraphics[width=0.35\columnwidth]{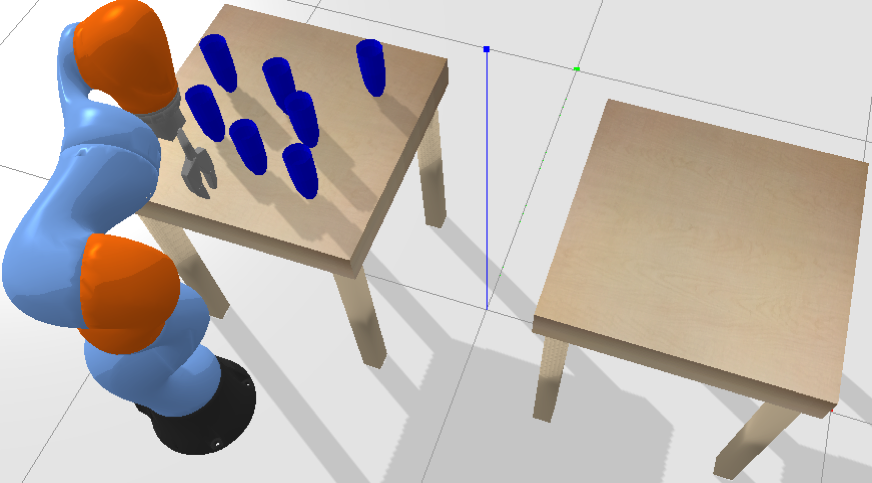}
    \includegraphics[width=0.35\columnwidth]{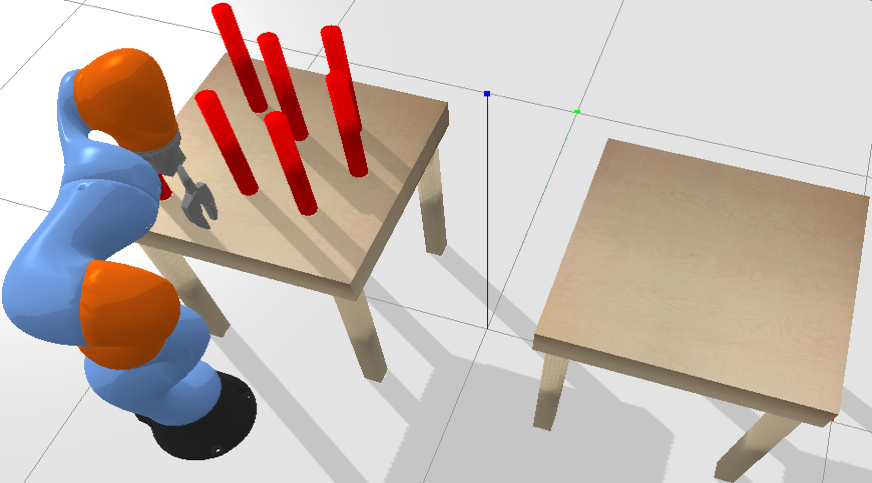}
    \includegraphics[width=0.35\columnwidth]{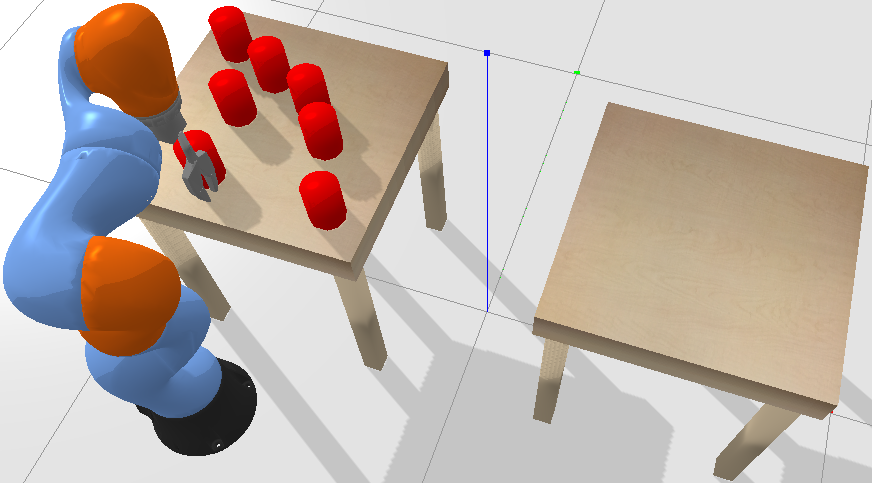}
    \caption{We address the problem of learning values for continuous
      action parameters in task and motion planning. We take a
      \emph{modular meta-learning} approach, where we train functions
      across multiple tasks to generate these values, by interleaving
      planning and optimization. Pictured are three of our tabletop
      manipulation tasks used for training and one used for evaluation
      (bottom right). Object shapes and sizes vary across tasks,
      affecting how the robot should choose grasp and place poses. On
      an evaluation task, the robot is given a small amount of data --
      it must \emph{learn quickly to plan quickly}, adapting its
      learned functions to generate good target poses.}
  \label{fig:tasks}
\end{figure}

We compare two different \emph{modular meta-learning} strategies: one,
based on \maml{}~\cite{maml}, focuses on learning neural network
weights that can be quickly adapted via gradient descent in a new
task; the other, based on \bouncegrad{}~\cite{moma}, focuses on
learning a fixed set of neural network ``modules'' from which we can
quickly choose a subset for a new task.

We demonstrate the effectiveness of these approaches in an object
manipulation domain, illustrated in \figref{fig:tasks}, in which the
robot must move all of the objects from one table to another.  This
general goal is constant across tasks, but different tasks will vary
the object shapes and sizes, requiring the robot to learn to
manipulate these different types of objects. We conjecture that the
meta-learning approach will allow the robot, at meta-training time, to
discover generally useful, task-independent strategies, such as
placing objects near the back of the available space; and, at
deployment time, to quickly learn to adapt to unseen object
geometries. Our methods are agnostic to exactly which aspects of the
environments are common to all tasks and which vary -- these concepts
are naturally discovered by the meta-learning algorithm.  We show that
the meta-learning strategies perform better than both a random sampler
and a reasonable set of hand-built, task-agnostic, uninformed
\spec{}s.

\section{Related Work}
\label{sec:relatedwork}
Our work is focused on choosing continuous action parameters within
the context of a symbolic plan.  We are not learning control policies
for
tasks~\cite{peters2013towards,kober2013reinforcement,ArgallReview10},
nor are we learning planning models of tasks~\cite{PasulaJAIR07}. We
assume any necessary policies and planning models exist; our goal is
to make planning with such models more efficient by learning \spec{}s
via modular meta-learning. There is existing work on learning samplers
for continuous action parameters in task and motion
planning~\cite{ltampkim1,ltampkim2,ltampwang,ltampchitnis}, but these
methods do not explicitly consider the problem of learning samplers that
can be quickly adapted to new tasks.

\emph{Meta-Learning:} Meta-learning is a particularly important
paradigm for learning in robotics, where training data can be very
expensive to acquire, because it dramatically lowers the data
requirements for learning new tasks.  Although it has a long history
in the transfer learning literature~\cite{pan2010survey},
meta-learning was recently applied with great effectiveness to
problems in robotics by Finn et al.~\cite{maml}.

\emph{Learning Modular Structure:} Our approach is a {\em modular}
learning approach, in the sense of Andreas et
al.~\cite{andreas2016neural}: the \spec{}s we learn are
associated with planning operators, allowing them to be recombined in
new ways to solve novel problems.  Andreas et al.~\cite{policysketch}
use reinforcement learning to train subtask modules in domains with
decomposable goals. Unlike in our work, they assume a policy sketch
is given.

\emph{Modular Meta-Learning:} Modular meta-learning was developed by
Alet et al.~\cite{moma} and forms the inspiration for this work. Their
work includes an EM-like training procedure that alternates between
composing and optimizing neural network modules, and also includes a
mechanism for choosing the best compositional structure of the modules
to fit a small amount of training data on a new task.

\section{Problem Setting}
\label{sec:problemsetting}

\subsection{Task and Motion Planning}
\label{subsec:tamp}
Robot task and motion (\tamp{}) problems are typically formulated as
discrete-time planning problems in a hybrid discrete-continuous state
transition system~\cite{GarrettIJRR,tamplgp}, with discrete variables
modeling which objects are being manipulated and other task-level
aspects of the domain, and continuous variables modeling the robot
configuration, object poses, and other continuous properties of the
world.

A {\em world state} $s \in \St$ is a complete description of the state
of the world, consisting of $(c, o^1, \ldots, o^n, x^1, \ldots, x^m)$,
where $c$ is the robot configuration, the $o^i$ are the states of each
object in the domain, and the $x^i$ are other discrete or continuous
state variables. Each object's state is $d$-dimensional and describes
its properties, such as mass or color.

We now define a \tamp{} problem, using some definitions from Garrett
et al.~\cite{GarrettIJRR}.  A {\em predicate} is a Boolean function.
A {\em fluent} is an evaluation of a predicate on a tuple
$(\bar{o}, \bar{\theta})$, where $\bar{o}$ is a set of discrete
objects and $\bar{\theta}$ is a set of continuous values. A set of
{\em basis predicates} $B$ can be used to completely describe a world
state. Given a world state $s$, the set of {\em basis fluents}
$\Phi_B(s)$ is the maximal set of atoms that are true in $s$ and can
be constructed from the basis predicate set $B$.  A set of {\em
  derived predicates} can be defined in terms of basis predicates. A
{\em planning state} $\I$ is a set of fluents, including a complete
set of basis fluents and any number of derived fluents; it is assumed
that any fluent not in $\I$ is false.

An {\em action} or \emph{operator} $a$ is given by an argument tuple
$\bar{O} = (O_1,\ldots, O_v)$, a parameter tuple
$\bar{\Theta} = (\Theta_1, \ldots, \Theta_w)$, a set of fluent
preconditions ${\it pre}(a)$ on $(\bar{O}, \bar{\Theta})$, and a set
of positive and negative fluent effects ${\it eff}(a)$ on
$(\bar{O}, \bar{\Theta})$.  An {\em action instance}
$a(\bar{o}, \bar{\theta})$ is an action $a$ with arguments and
parameters $(\bar{O}, \bar{\Theta})$ replaced with discrete objects
$\bar{o}$ and continuous values $\bar{\theta}$.  An action instance
$a(\bar{o},\bar{\theta})$ is {\em applicable} in planning state $\I$
if ${\it pre}(a(\bar{o},\bar{\theta})) \subseteq \I$.  The result of
{\em applying} an action instance $a(\bar{o}, \bar{\theta})$ to a
planning state $\I$ is a new state
$\mathcal I \cup {\it eff}^+(a(\bar{o},\bar{\theta})) \setminus {\it
  eff}^-(a(\bar{o},\bar{\theta}))$, where ${\it eff}^+$ and
${\it eff}^-$ are the positive and negative fluents in ${\it eff}$,
respectively.  For $a$ to be a well-formed action, ${\it eff}^+(a)$
and ${\it eff}^-(a)$ must be structured so that the planning state
${\mathcal I}'$ resulting from applying $a$ is valid (contains a
complete set of basis fluents). Then, $\mathcal{I'}$ determines a world
state $s \in \St$, and the action $a$ can be viewed as a deterministic
transition on world states.

A \tamp{} problem $({\mathcal A}, \I, {\mathcal G})$ is given by a set
of actions ${\mathcal A}$, an initial planning state $\I$, and a goal
set of fluents ${\mathcal G}$. A sequence of action instances,
$\pi = \langle a_1(\bar{o}_1, \bar{\theta}_1), \ldots a_k(\bar{o}_k,
\bar{\theta}_k)\rangle$, is called a \emph{plan}. A plan $\pi$ is a
{\em task-level solution} for problem
$({\mathcal A}, \I, {\mathcal G})$ if $a_1(\bar{o}_1, \bar{\theta}_1)$
is applicable in $\I$, each $a_i(\bar{o}_i, \bar{\theta}_i)$ is
applicable in the $(i-1)$th state resulting from application of the
previous actions, and ${\mathcal G}$ is a subset of the final state. A
\emph{plan skeleton} $\hat{\pi}$ is a sequence of actions whose
discrete arguments are instantiated but continuous parameters are not:
$\hat{\pi} = \langle a_1(\bar{o}_1, \bar{\Theta}_1), \ldots
a_k(\bar{o}_k, \bar{\Theta}_k)\rangle$.

A {\em world-state trajectory} $\tau(\pi, \I)$ is the sequence of
world states $s_1, \ldots, s_k \in \St$ induced by the sequence of
planning states $\I_1, ..., \I_k$ resulting from applying plan $\pi$ starting from
$\I$. A task-level solution $\pi$ is a {\em complete solution} for the
\tamp{} problem $({\mathcal A}, \I, {\mathcal G})$ if, letting
$\tau(\pi, \I) = \langle s_1, \ldots s_k \rangle$, there exist robot
trajectories $\tilde{\tau}_1, \ldots, \tilde{\tau}_{k-1}$ such that
$\tilde{\tau}_i$ is a collision-free path (a \emph{motion plan}) from
$c_{i}$ to $c_{i+1}$, the robot configurations in world states $s_{i}$
and $s_{i+1}$ respectively.

\subsection{Solving Task and Motion Planning Problems}
\label{subsec:solvingtamp}
Finding good search strategies for solving \tamp{} problems is an
active area of research, partly owing to the difficulty of finding
good continuous parameter values that produce a complete solution
$\pi$.  Our meta-learning method could be adapted for use in many
\tamp{} systems, but for clarity we focus on the simple one
sketched below.\footnote{Generally, not all of the elements of
  $\bar{\theta}$ are actually free parameters given a skeleton. Some
  elements of $\bar{\theta}$ may be uniquely determined by the
  discrete arguments or other parameters, or by the world state. We
  will not complicate our notation by explicitly handling these
  cases.} See \secref{sec:experiments} for discussion of the \tamp{}
system we use in our implementation.

\removelatexerror
\begin{algorithm}[H]
  \SetAlgoNoEnd
  \DontPrintSemicolon
  \SetKwFunction{algo}{algo}\SetKwFunction{proc}{proc}
  \SetKwProg{myalg}{Algorithm}{}{}
  \SetKwProg{myproc}{Subroutine}{}{}
  \SetKw{Continue}{continue}
  \SetKw{Break}{break}
  \SetKw{Return}{return}
  \myalg{\textsc{PlanSketch}$(\A, \I, \G)$}{
    \nl \For{$\hat{\pi}$ in some enumeration of plan skeletons}{
    \nl \For{$\bar{\theta}$ s.t. $\pi = \hat{\pi}(\bar{\theta})$
      is a task-level solution}{
    \nl \If{motion plans exist for $\pi$}{
       \Return{the complete solution $\pi$}}}}}
\label{alg:planSketch}
\end{algorithm}

The problems of symbolic task planning to yield plausible plan
skeletons (Line 1) and collision-free motion planning (Line 3) are
both well-studied, and effective solutions exist. We focus on the
problem of searching over continuous values $\bar{\theta}$ for the
skeleton parameters (Line 2).

We first outline two simple strategies for finding
$\bar{\theta}$.  In {\em random sampling}, we perform a simple
depth-first backtracking search: sample values for $\theta_i$
uniformly at random from some space, check whether there is a motion
plan from $s_{i-1}$ to $s_{i}$, continue on to sample $\theta_{i+1}$
if so, and either sample $\theta_i$ again or backtrack to a higher
node in the search tree if not.  In the {\em hand-crafted} strategy,
we rely on a human programmer to write one or more \emph{\spec{}s}
$\sigma_a^i$ for each action $a$.  A \spec{} is a function
$\sigma(\I, \bar{o}, j)$, where $\I$ is a planning state, $\bar{o}$
are the discrete object arguments with which $a$ will be applied, and
$j$ is the step of the skeleton where a particular instance of $a$
occurs. The specializer $\sigma$ returns a vector of continuous
parameter values $\bar{\theta}$ for $a$.  So, in this hand-crafted
strategy, for each plan skeleton $\hat{\pi}$ we need only consider the
following discrete set of plans $\pi$:
\[\pi=\langle a_1(\bar{o}_1, \sigma^{i_1}_{a_1}(\I_{1}, \bar{o}_1, 1)), \ldots
  a_k(\bar{o}_k, \sigma^{i_k}_{a_k}(\I_{k}, \bar{o}_k, k))
  \rangle\;\;,\] where the $i$ values select which specializer to use
for each step. Each setting of the $i$ values yields a different plan
$\pi$.

Now, it is sensible to combine the search over both skeletons and
\spec{}s into a single discrete search. Let $\Sigma(W)$ be a set of
\spec{}s (the reason for this notation will become clear in the next
section) and $\A(\Sigma(W))$ be a discrete set of ``actions'' obtained
by combining each action $a \in \A$ with each specializer
$\sigma_a^i \in \Sigma(W)$, indexed by $i$. We obtain our algorithm
for planning with \spec{}s: \removelatexerror
\begin{algorithm}[H]
  \SetAlgoNoEnd
  \DontPrintSemicolon
  \SetKwFunction{algo}{algo}\SetKwFunction{proc}{proc}
  \SetKwProg{myalg}{Algorithm}{}{}
  \SetKwProg{myproc}{Subroutine}{}{}
  \SetKw{Continue}{continue}
  \SetKw{Break}{break}
  \SetKw{Return}{return}
  \myalg{\textsc{Plan}$(\A, \I, \G, \Sigma(W))$}{
    \nl \For{$\pi$ in $\textsc{SymbolicPlan}(\A(\Sigma(W)), \I, \G)$}{
      \nl \If{motion plans exist for $\pi$}{
        \Return{the complete solution $\pi$}}}}
\label{alg:plan}
\end{algorithm}

\subsection{Learning \Spec{}s}
We begin by defining our learning problem for just a single task. A
single-task {\em \spec{} learning} problem is a tuple
$({\mathcal A}, \data, \Sigma(W))$, where ${\mathcal A}$ is a set of
actions specifying the dynamics of the domain,
$\data = \{(\I_1, {\mathcal G}_1), \ldots, (\I_n, {\mathcal G}_n)\}$
is a dataset of (initial state, goal) problem instances,
$\Sigma$ is a set of functional forms for \spec{}s (such as neural
network architectures), and $W$ is a set of initial weights such that
$\Sigma(W)$ is a set of fully instantiated \spec{}s that can be used
for planning with the \textsc{Plan} algorithm.

The objective of our learning problem is to find $W$ such that
planning with $\Sigma(W)$ will, in expectation over new
$(\I, {\mathcal G})$ problem instances drawn from the same
distribution as $\data$, be likely to generate complete solutions. The
functional forms $\Sigma$ of the specializers are given (just like in
the hand-crafted strategy), but the weights $W$ are learned.

Although our ultimate objective is to improve the efficiency of the
overall planner, that is done by replacing the search over continuous
parameter values $\bar{\theta}$ with a deterministic choice or search
over a finite set of parameter values provided by the \spec{}s; so, our
objective is really that these \spec{}s be able to solve problems from
$\data$.

Most directly, we could try to minimize $0-1$ {\em single-task loss}
on $\data$, so that $W^* = \argmin_W \Loss_S(W; \data)$ where:
\[\Loss_S(W; \data) = \sum_{(\I,\G)\in\data} \begin{cases}
0 & \text{$\procm{Plan}(\A, \I, \G, \Sigma(W))$ succeeds}\\
1 & \text{otherwise}
\end{cases}
\label{eq:idealLoss}
\]
Unfortunately, this loss is much too difficult to optimize in
practice; in \secref{subsec:learn} we will outline strategies for
smoothing and approximating the objective.

\subsection{Meta-Learning \Spec{}s}
In meta-learning, we wish to learn, from several training tasks, some
form of a prior that will enable us to learn to perform well quickly
on a new task. A {\em \spec{} meta-learning} problem, given by a tuple
$({\mathcal A}, (\data_1, \ldots, \data_m), \Sigma(W))$, differs from
a single-task specializer learning problem both in that it has
multiple datasets $\data_j$, and in that it has a different objective.
We make the implicit assumption, standard in meta-learning settings,
that there is a hierarchical distribution over $(\I, {\mathcal G})$
problems that the robot will encounter: we define a {\em task} to be a
single distribution over $(\I, \G)$, and we assume we have a
distribution over tasks.

Let $\procm{Learn}({\mathcal A}, \data, \Sigma(W))$ be a specializer
learning algorithm that returns $W^*$, tailored to work well on
problems drawn from $\data$. Our meta-learning objective will then be
to find a value of $W$ that serves as a good prior for {\sc Learn} on
new tasks, defined by new $(\I, {\mathcal G})$
distributions. Formally, the meta-learning objective is to find
$W^*_M = \argmin_W \Loss_M(W)$, where the meta-learning loss is, letting
$j$ index over tasks:
\[
  \Loss_M(W) = \frac{1}{m} \sum_{j=1}^m \Loss_S(\procm{Learn}(\A, \data_j^{\rm
    train}, \Sigma(W)); \data_j^{\rm test})\;\;.
\]
The idea is that a new set of weights obtained by starting with $W$
and applying \procm{Learn} on a training set from task $j$ should
perform well on a held-out test set from task $j$.

After learning $W^*_M$, the robot is deployed. When it is given a
small amount of training data $\data_{\rm new}$ drawn from a new task,
it will call $\procm{Learn}(\A, \data_{\rm new}, \Sigma(W^*_M))$ to
get a new set of weights $W^*_{\rm new}$, then use the planner
$\procm{Plan}(\A, \I, \G, \Sigma(W^*_{\rm new}))$ to solve future
problem instances $(\A, \I, \G)$ from this new task. If the
meta-learning algorithm is effective, it will have \emph{learned} to
\begin{itemize}
\item learn quickly (from a small dataset $\data_{\rm new}$) to
\item plan quickly (using the \spec{}s $\Sigma(W^*_{\rm new})$ in
  place of a full search over continuous parameter values
  $\bar{\theta}$),
\end{itemize}
motivating our title.

\section{Algorithms}
\label{sec:approach}
In this section, we begin by describing two single-task specializer
learning algorithms, and then we discuss a specializer meta-learning
algorithm that can be used with any specializer learning algorithm.

\subsection{Single-Task Specializer Learning Algorithms}
\label{subsec:learn}
Recall that an algorithm for learning \spec{}s takes as input
$({\mathcal A}, \data, \Sigma(W))$, and its job is to return
$W^* = \argmin_W \Loss_S(W; \data)$. We consider two algorithms: {\em
  alternating descent} (\ad) and {\em subset selection} (\subs).

\emph{a) Alternating Descent:} \ad{} adjusts the weights $W$ to
tune them to dataset $\data$.

If we knew, for each $({\mathcal I}, {\mathcal G}) \in \data$, the
optimal plan skeleton and choices of \spec{}s that lead to a complete
solution $\pi$ for the \tamp{} problem
$({\mathcal A}, {\mathcal I}, {\mathcal G})$, then we could adjust the
elements of $W$ corresponding to the chosen specializers in order to
improve the quality of $\pi$. However, this optimal set of actions and
\spec{}s is not known, so we instead perform an EM-like {\em
  alternating optimization}. In particular, we use the {\sc PlanT}
algorithm (described in detail later), an approximation of {\sc Plan}
that can return illegal plans, to find a skeleton $\hat{\pi}$ and
sequence of specializers $\sigma_j$ to be optimized. Then, we adjust
the elements of $W$ corresponding to the $\sigma_j$ to make the plan
``less illegal.''

Formally, we assume the existence of a {\em predicate loss
  function} $\Loss_p$ for each predicate $p$ in the domain, which takes in
the arguments of predicate $p$ ($\bar{o}$ and $\bar{\theta}$) and a
world state $s \in \St$, and returns a positive-valued loss measuring
the degree to which the fluent $p(\bar{o}, \bar{\theta})$ is violated
in $s$. If $p(\bar{o}, \bar{\theta})$ is true in $s$, then
$\Loss_p(\bar{o}, \bar{\theta}, s)$ must be zero.  For example, if fluent
$\phi = {\rm pose}(o, v)$ asserts that the pose of object $o$ should
be the value $v$, then we might use the squared distance $(v - v')^2$
as the predicate loss, where $v'$ is the actual pose of $o$ in $s$.

Now consider the situation in which we run {\sc PlanT}, and it returns a
plan $\pi$ created from a plan skeleton $\hat{\pi}$ and the chosen
\spec{}s $\sigma_1, \ldots, \sigma_k$. From this, we can compute both
the induced trajectory of planning states
${\mathcal I_1}, \ldots, {\mathcal I_k}$, and the induced trajectory
of world states $\tau = \langle s_1, \ldots s_k \rangle$.  We can now
define a {\em trajectory loss} function $\Loss_\tau$ on $W$ for $\pi$:
\[\Loss_\tau(W; \I, \G, \pi) = \sum_{j = 1}^k \sum_{\phi \in {\it
      eff}^+(a_j(\bar{o}_j, \bar{\theta}_j))} \Loss_{p(\phi)}(\bar{o}_j,
     \bar{\theta}_j, s_j)\;\;.
\]
This is a sum over steps $j$ in the plan, and for each step, a sum
over positive fluents $\phi$ in its effects, of the degree to which
that fluent is violated in the resulting world state $s_j$. Here,
$p(\phi)$ is the predicate associated with fluent $\phi$.  Recall that
$\bar{\theta}_j = \sigma_j(\I_j, \bar{o}_j, j; W)$, where we have
included $W$ to expose the \spec{}s' parametric forms. Thus, we have:
\[\Loss_\tau(W; \I, \G, \pi) = \sum_{j=1}^k \sum_{\phi} \Loss_{p(\phi)}(\bar{o}_j,
     \sigma_j(\I_j, \bar{o}_j, j; W), s_j)\;\;.
   \]

   If the $\Loss_p$ are differentiable with respect to the
   $\bar{\theta}$, and the functional forms $\Sigma$ generating the
   $\bar{\theta}$ are differentiable with respect to $W$ and their
   continuous inputs, then $W$ can be adjusted via a gradient step to
   reduce $\Loss_\tau$. This method will adjust only the values of $W$
   that were used in the \spec{}s chosen by {\sc PlanT}.
   The overall algorithm is: \removelatexerror
\begin{algorithm}[H]
  \SetAlgoLined
  \SetAlgoNoEnd
  \DontPrintSemicolon
  \SetKwFunction{algo}{algo}\SetKwFunction{proc}{proc}
  \SetKwProg{myalg}{Algorithm}{}{}
  \SetKwProg{myproc}{Subroutine}{}{}
  \SetKw{Continue}{continue}
  \SetKw{Break}{break}
  \SetKw{Return}{return}
  \myalg{\textsc{AD-Learn}$(\A, \data, \Sigma(W), n_{\rm iters},
    n_{\rm plans})$}{
    \nl \For{$t = 1$ to $n_{\rm iters}$}{
      \nl Sample $(\I, \G)$ from $\data$. \;
      \nl $\pi \gets$ \textsc{PlanT}$(\A, \I, \G, \Sigma(W), t, n_{\rm
        plans})$ \;
      \nl $W \gets W - \alpha \nabla_{W}\Loss_\tau(W; \I, \G, \pi)$ \;
      }}
  \myproc{\textsc{PlanT}$(\A, \I, \G, \Sigma(W), t, n_{\rm plans})$}{
    \nl \For{$i = 1$ to $n_{\rm plans}$}{
    \nl $\pi_i \gets {\rm next}\; \textsc{SymbolicPlan}(\A(\Sigma(W)), \I, \G)$\;
    \nl \If{motion plans exist for $\pi_i$}{
    \nl ${\rm score}(\pi_i) \gets -\Loss_\tau(W; \I, \G, \pi_i)$}}
    \nl \If{\rm no scores were computed}{
    \nl      Randomly initialize more \spec{}s; repeat.}
    \nl \Return{\rm sample $\pi_i \sim e^{{\rm score}(\pi_i)/T(t)} / Z$}}
\label{alg:adLearn}
\end{algorithm}

We now describe in detail the $\procm{PlanT}$ procedure, which is an
approximation of $\procm{Plan}$.  Generally, while we are learning
$W$, we will not have a complete and correct set of specializers, but
we still need to assemble plans in order to adjust the $W$. In
addition, to prevent local optima, and inspired by the use of
simulated annealing for structure search in \bouncegrad{}~\cite{moma}
and \moma{}~\cite{moma}, we do not always consider the $\pi$ with
least loss early on. {\sc PlanT}, rather than trying to find a $\pi$
that is a complete solution for the \tamp{} problem, treats {\sc SymbolicPlan}
as a generator, generates $n_{\rm plans}$ symbolic plans that are not
necessarily valid solutions, and for each one that is feasible with
respect to motion planning, computes its loss.  It then samples a plan
to return using a Boltzmann distribution derived from the losses, with
``temperature'' parameter $T(t)$ computed as a function of the number
of optimization steps done so far. This $T(t)$ should be chosen to go
to zero as $t$ increases.

\emph{b) Subset Selection:} \subs{} assumes that $\Sigma(W)$ includes
a large set of \spec{}s, and simply selects a subset of them to use
during planning, without making any adjustments to the weights $W$.
Let $\Sigma_a(W)$ be the set of \spec{}s for action $a$ and integer
$k$ be a parameter of the algorithm.  The \subs{} algorithm simply
finds, for each action $a$, the size-$k$ subset $\rho_a$ of
$\Sigma_a(W)$ such that $\Loss_S(\cup_a \rho_a; \data)$ is
minimized\footnote{Technically speaking, the first argument to
  $\Loss_S$ should be all the weights $W$; we can assume that
  $\cup_a \rho_a$ is the following operation: leave the elements of
  $W$ that parameterize the $\rho_a$ unchanged, and set the rest to
  0.}. There are many strategies for finding such a set; in our
experiments, we have a small number of actions and set $k = 1$, and so
we can exhaustively evaluate all possible combinations.

\subsection{Specializer Meta-Learning Algorithm}
Recall that an algorithm for meta-learning \spec{}s takes as input
$({\mathcal A}, (\data_1, \ldots, \data_m), \Sigma(W))$, and its job
is to return $W^*_M = \argmin_W \Loss_M(W)$, which should be a good
starting point for learning in a new task. This ideal objective is
difficult to optimize, so we must make approximations.

We begin by describing the meta-learning algorithm, which follows a
strategy very similar to \maml{}~\cite{maml}. We do stochastic gradient descent in
an outer loop: draw a task $\data_j$ from the task distribution, use
some learning algorithm $\procm{Learn}$ to compute a new set of
weights $W_j$ for $\data_j^{\rm train}$ starting from $W$, and update
$W$ with a gradient step to reduce the trajectory loss on
$D_j^{\rm test}$ evaluated using $W_j$. \removelatexerror
\begin{algorithm}[H]
  \SetAlgoLined
  \SetAlgoNoEnd
  \DontPrintSemicolon
  \SetKwFunction{algo}{algo}\SetKwFunction{proc}{proc}
  \SetKwProg{myalg}{Algorithm}{}{}
  \SetKwProg{myproc}{Subroutine}{}{}
  \SetKw{Continue}{continue}
  \SetKw{Break}{break}
  \SetKw{Return}{return}
  \myalg{\textsc{MetaLearn}$(\A, \data_1, \ldots, \data_m, \Sigma(W))$}{
    \nl \While{not done}{
    \nl $j \gets {\rm sample}(1, \ldots, m)$\;
    \nl $W_j \gets \procm{Learn}(\A, \data_j^{\rm train}, \Sigma(W))$\;
    \nl $W \gets W - \beta \nabla_{W} \Loss_{\tau, \data_j^{\rm test}}(\Sigma(W_j))$\;
       }}
\label{alg:metalearn}
\end{algorithm}
For efficiency, in practice we drop the Hessian term in the gradient
by taking the gradient with respect to $W_j$ (so
$\nabla_{W} \to \nabla_{W_j}$). This is an approximation that is
successfully made by several \maml{} implementations. We define:

{\small
\[\Loss_{\tau, \data}(\Sigma(W)) =
  \sum_{\I, \G \in \data} \Loss_\tau(W; \I, \G,
  \procm{PlanT}(\A,\I,\G,\Sigma(W),\infty,\infty)).\]} This
expression gives the smoothed trajectory loss for the best plan we can
find using the given $\Sigma(W)$, summed over all planning problems in
$\data$.  When we compute the gradient, we ignore the dependence of
the plan structure on $W$.  Thus, we estimate
$\nabla_{W_j} \Loss_{\tau, \data_j^{\rm test}}(\Sigma(W_j))$ as
follows:
\begin{algorithm}[H]
  \SetAlgoLined
  \SetAlgoNoEnd
  \DontPrintSemicolon
  \SetKwFunction{algo}{algo}\SetKwFunction{proc}{proc}
  \SetKwProg{myalg}{Algorithm}{}{}
  \SetKwProg{myproc}{Subroutine}{}{}
  \SetKw{Continue}{continue}
  \SetKw{Break}{break}
  \SetKw{Return}{return}
    \nl \For{$t = 1$ to $n_{\rm gradEst}$}{
      \nl Sample $(\I, \G)$ from $\data_j^{\rm test}$. \;
      \nl $\pi \gets$ \textsc{PlanT}$(\A, \I, \G, \Sigma(W_j), \infty, \infty)$ \;
      \nl $\nabla_{W_j} \gets \nabla_{W_j} + \nabla_{W_j}\Loss_\tau(W_j;
      \I, \G, \pi)$ \;
      }
    \nl \Return{$\nabla_{W_j}$}
  \end{algorithm}

  When $\procm{Learn}$ is the subset selection learner (\subs{}), the
  {\sc Learn} procedure returns only a subset of the $\Sigma(W)$,
  corresponding to the chosen specializers. Only the weights in that
  subset are updated with a gradient step on the test data.

\section{Experiments}
\label{sec:experiments}

We demonstrate the effectiveness of our approach in a simulated object
manipulation domain where the robot is tasked with moving all
objects from one table to another, as shown in~\figref{fig:tasks}. The
object geometries vary across tasks, while a single task is a
distribution over initial configurations of the objects on the
starting table. We consider 6 training tasks and 3 evaluation tasks,
across 3 object types: cylinders, bowls, and vases. The phrase ``final
task'' henceforth refers to a random sample of one of the 3 evaluation
tasks.

We use a KUKA iiwa robot arm. Grasp legality is computed using a
simple end effector pose test based on the geometry of the object
being grasped. We require that cylinders are grasped from the side,
while bowls and vases are grasped from above, on their lip.  There are
four operators: \texttt{moveToGrasp} and \texttt{moveToPlace} move the
robot (and any held object) to a configuration suitable for grasping
or placing an object, \texttt{grasp} picks an object up, and
\texttt{place} places it onto the table. All operators take in the ID
of the object being manipulated as a discrete argument. The
continuous parameters learned by our specializers are the target end
effector poses for each operator; we use an inverse kinematics solver
to try reaching these poses.

We learn three specializers for each of the first three operators, and
one specializer for \texttt{place} due to its relative simplicity. The
state representation is a vector containing the end effector pose,
each object's position, object geometry information, robot base
position, and the ID of the currently-grasped object (if any). Thus,
we are assuming a fully observed closed world with known object poses,
in order to focus on the meta-learning aspect of this setting.

All specializers are fully connected, feedforward neural networks with
hidden layer sizes [100, 50, 20], a capacity which preliminary
experiments found necessary. We use batch size 32 and the Adam
optimizer~\cite{adam} with initial learning rate $10^{-2}$, decaying
by 10\% every 1000 iterations.

For motion planning, we use the RRT-Connect
algorithm~\cite{rrtconnect}; we check for infeasibility crudely by
giving the algorithm a computation allotment, implemented as a maximum
number of random restarts to perform, upon which a (infeasible)
straight-line trajectory is returned. We use
Fast-Forward~\cite{hoffmann2001ff} as our symbolic planner. For
simulation and visualization, we use the pybullet~\cite{pybullet}
software.

A major source of difficulty in this domain is that the end effector
poses chosen by the specializers must be consistent with both each
other (place pose depends on grasp pose, etc.) and the object
geometries. Furthermore, placing the first few objects near the front
of the goal table would impede the robot's ability to place the
remaining objects. We should expect that the general strategies
discovered by our meta-learning approach would handle these
difficulties.

To implement the discrete search over plan skeletons and specializers,
we adopt the \tamp{} approach of Srivastava et
al.~\cite{tampinterface}, which performs optimistic classical planning
using abstracted fluents, attempts to find a feasible motion plan, and
incorporates any infeasibilities back into the initial state as
logical fluents. For each skeleton, we search exhaustively over all
available specializers for each operator.

\emph{Evaluation:} We evaluate the {\sc MetaLearn} algorithm with both
the alternating descent (\ad) learner and the subset selection (\subs)
learner. We test against two baselines, random sampling and the
hand-crafted strategy, both of which are described
in~\secref{subsec:solvingtamp}. The random sampler is conditional, sampling
only end effector poses that satisfy the kinematic constraints of the
operators. At final task time with the \ad\ learner, we optimize the
specializers on 10 batches of training data, then evaluate on a test
set of 50 problems from this task. At final task time with the \subs\
learner, we choose a subset of $k = 1$ specializer per operator that
performs the best over one batch of training data, then use only that
subset to evaluate on a test set. Note that we should expect the test
set evaluation to be much faster with the \subs\ learner than with the
\ad\ learner, since we are planning with fewer specializers.

\begin{table}[t]
  \vspace{0.6em}
  \centering
  \resizebox{\columnwidth}{!}{
  \tabcolsep=0.08cm{
  \begin{tabular}{c|c|c|c|c|c}
    \toprule[1.5pt]
    \textbf{Setting} & \textbf{System} & \textbf{Final Task Solve \%} & \textbf{Train Iters to 50\%} & \textbf{Search Effort} & \textbf{Train Time (Hours)}\\
    \midrule[2pt]
    3 obj. & Baseline: Random & 24 & N/A & 52.2 & N/A\\
    \midrule
    3 obj. & Baseline: Hand-crafted & 68 & N/A & 12.1 & N/A\\
    \midrule
    3 obj. & Meta-learning: \ad & 100 & 500 & 2.5 & 4.3\\
    \midrule
    3 obj. & Meta-learning: \subs & 100 & 500 & 2.0 & 0.6\\
    \midrule[1.5pt]
    5 obj. & Baseline: Random & 14 & N/A & 81.3 & N/A\\
    \midrule
    5 obj. & Baseline: Hand-crafted & 44 & N/A & 34.3 & N/A\\
    \midrule
    5 obj. & Meta-learning: \ad & 88 & 2.1K & 8.6 & 7.4\\
    \midrule
    5 obj. & Meta-learning: \subs & 72 & 6.8K & 4.1 & 1.5\\
    \midrule[1.5pt]
    7 obj. & Baseline: Random & 0 & N/A & N/A & N/A\\
    \midrule
    7 obj. & Baseline: Hand-crafted & 18 & N/A & 64.0 & N/A\\
    \midrule
    7 obj. & Meta-learning: \ad & 76 & 5.1K & 18.3 & 12.3\\
    \midrule
    7 obj. & Meta-learning: \subs & 54 & 9.2K & 7.8 & 2.1\\
    \bottomrule[1.5pt]
  \end{tabular}}}
\caption{Summary of experimental results. Percentage of 50 final task
  problem instances solved within a 30-second timeout, number of
  meta-training iterations needed to reach 50\% final task solve rate,
  average number of plan skeletons and specializers searched over, and
  total training time in hours. Both meta-learning approaches learn to
  perform much better at the final task than the baselines
  do. Notably, the alternating descent (\ad) learner performs better
  than the subset selection (\subs) learner, likely because in the
  former, the specializer weights are optimized for the final task
  rather than held fixed. However, this improvement comes at the cost
  of much longer training times. Meta-learners were trained for
  $10^{4}$ iterations.}
\label{table:results}
\end{table}

\begin{figure}[t]
  \centering
  \vspace{0.75em}
    \noindent
    \includegraphics[width=0.49\columnwidth]{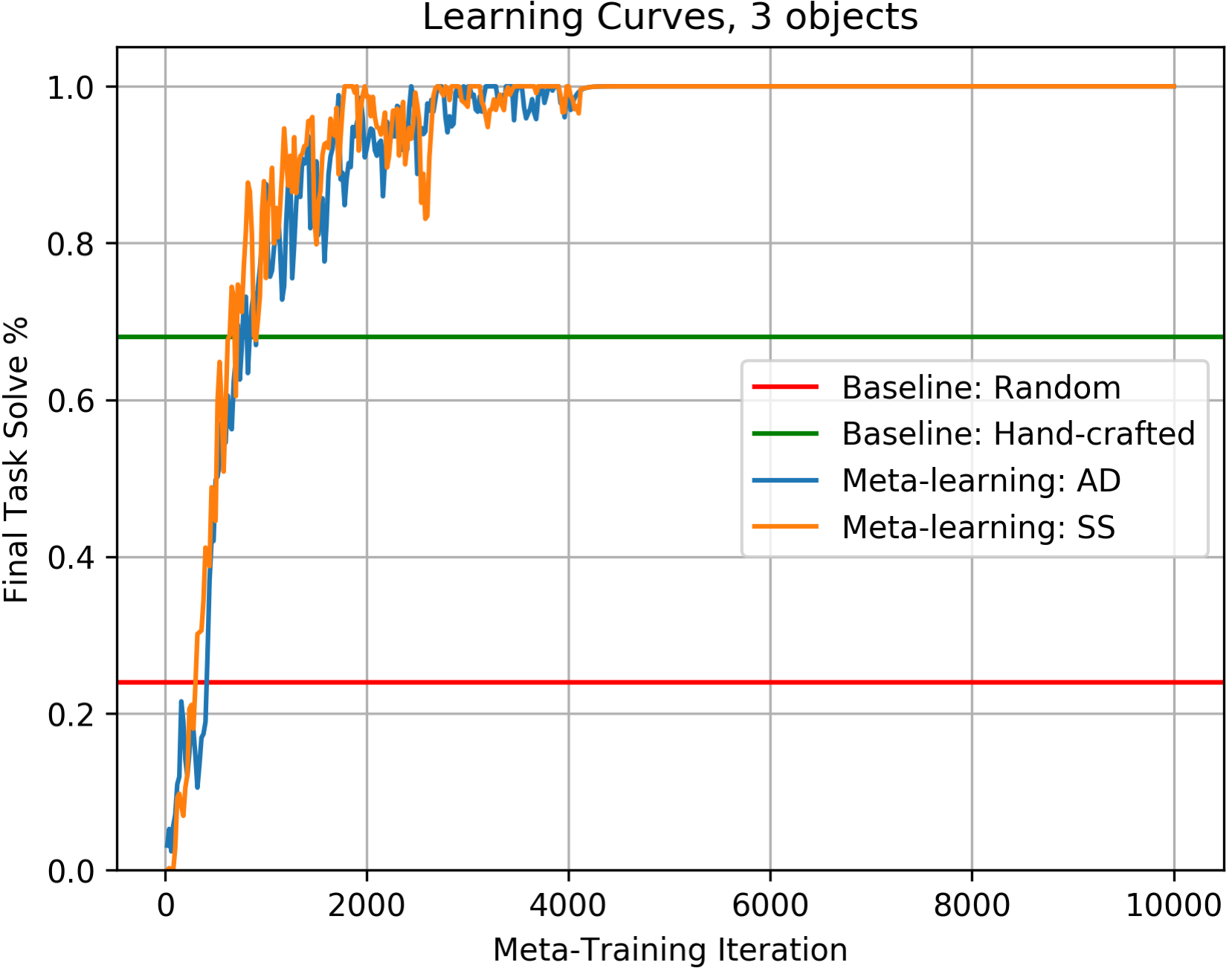}
    \includegraphics[width=0.49\columnwidth]{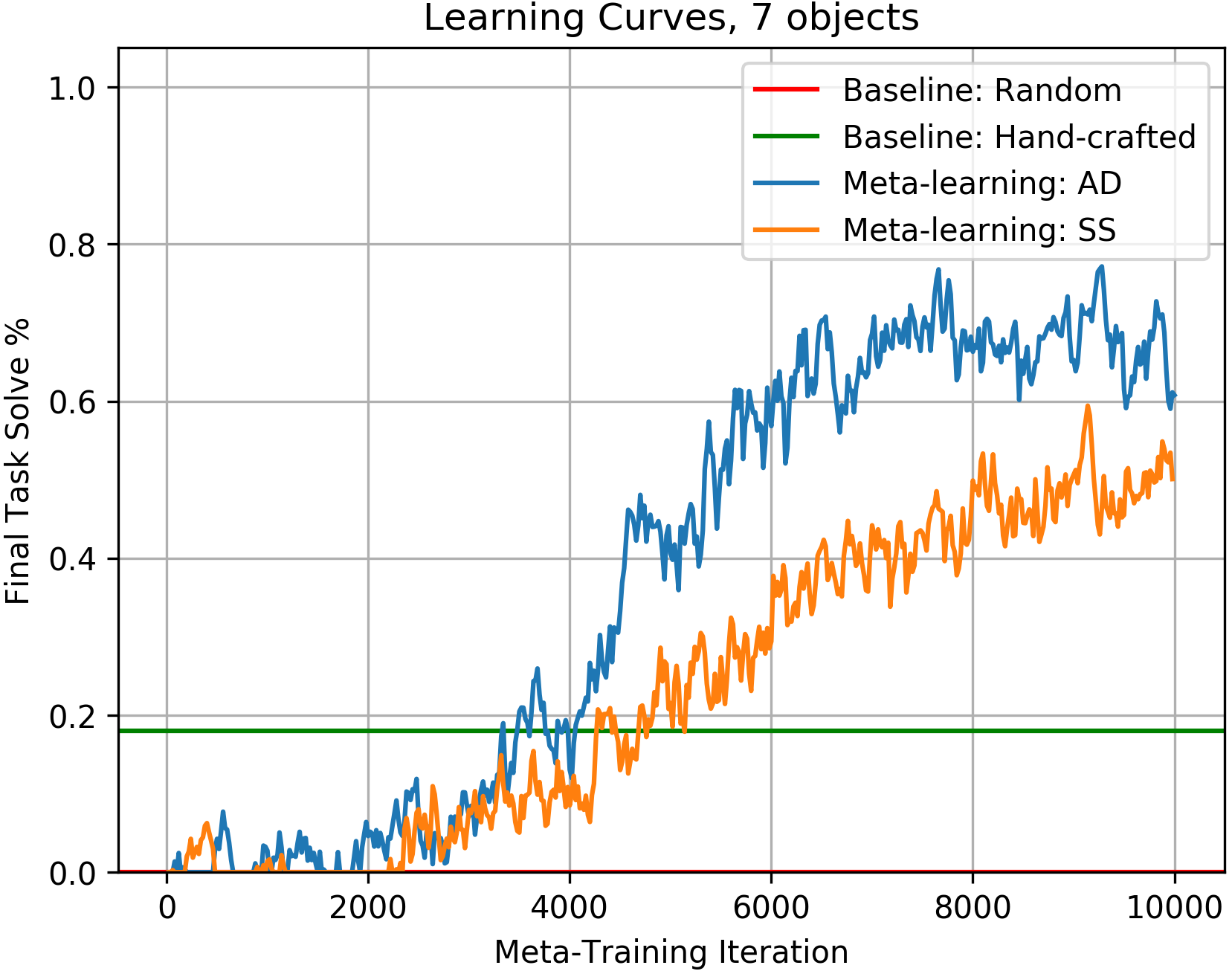}
    \caption{Learning curves over $10^{4}$ training iterations
      (smoothed).}
  \label{fig:learningcurves}
\end{figure}

\textbf{Results \& Discussion} \tabref{table:results} and
\figref{fig:learningcurves} show that both meta-learning approaches
perform much better at the final task than the baselines do. The
random sampler fails because it expends significant effort trying to
reach infeasible end effector poses, such as those behind the
objects. The hand-crafted specializers, though they perform better
than the random sampler, suffer from a lack of context: because they
are task-agnostic, they cannot specialize, and so search effort is
wasted on continuous parameter values that are inappropriate for the
current task, making timeouts frequent. Furthermore, the hand-crafted
strategy does not adapt to the state (e.g., the locations of objects
around one being grasped).

Qualitatively, we found that the general strategies we outlined
earlier for succeeding in this domain were meta-learned by our
approach (see video linked in abstract).

Notably, the alternating descent (\ad) learner performs better than
the subset selection (\subs) learner, likely because in the former,
the specializer weights are optimized for the final task rather than
held fixed. These findings suggest that this sort of fine-tuning is an
important step to learning specializers in this domain. However, this
improvement comes at the cost of much longer training times, since the
\ad\ learner performs an inner gradient computation which the \subs\
learner does not; the \ad\ learner may be impractical in larger
domains without introducing heuristics to guide the search. Another
finding is that the \subs\ learner expends much less search effort
than the \ad\ learner, as expected.

\begin{figure}[t]
  \centering
    \noindent
    \includegraphics[width=0.49\columnwidth]{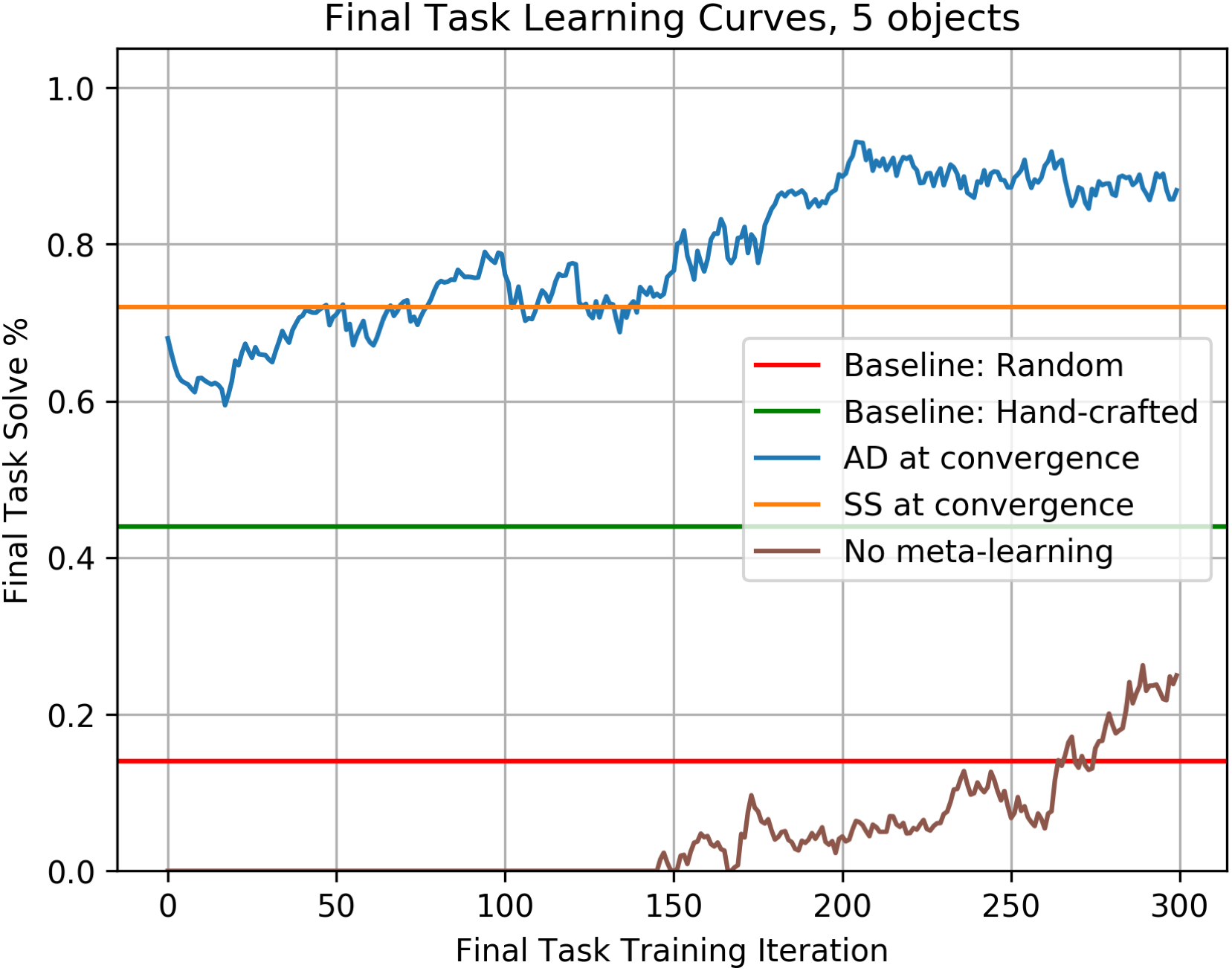}
    \includegraphics[width=0.49\columnwidth]{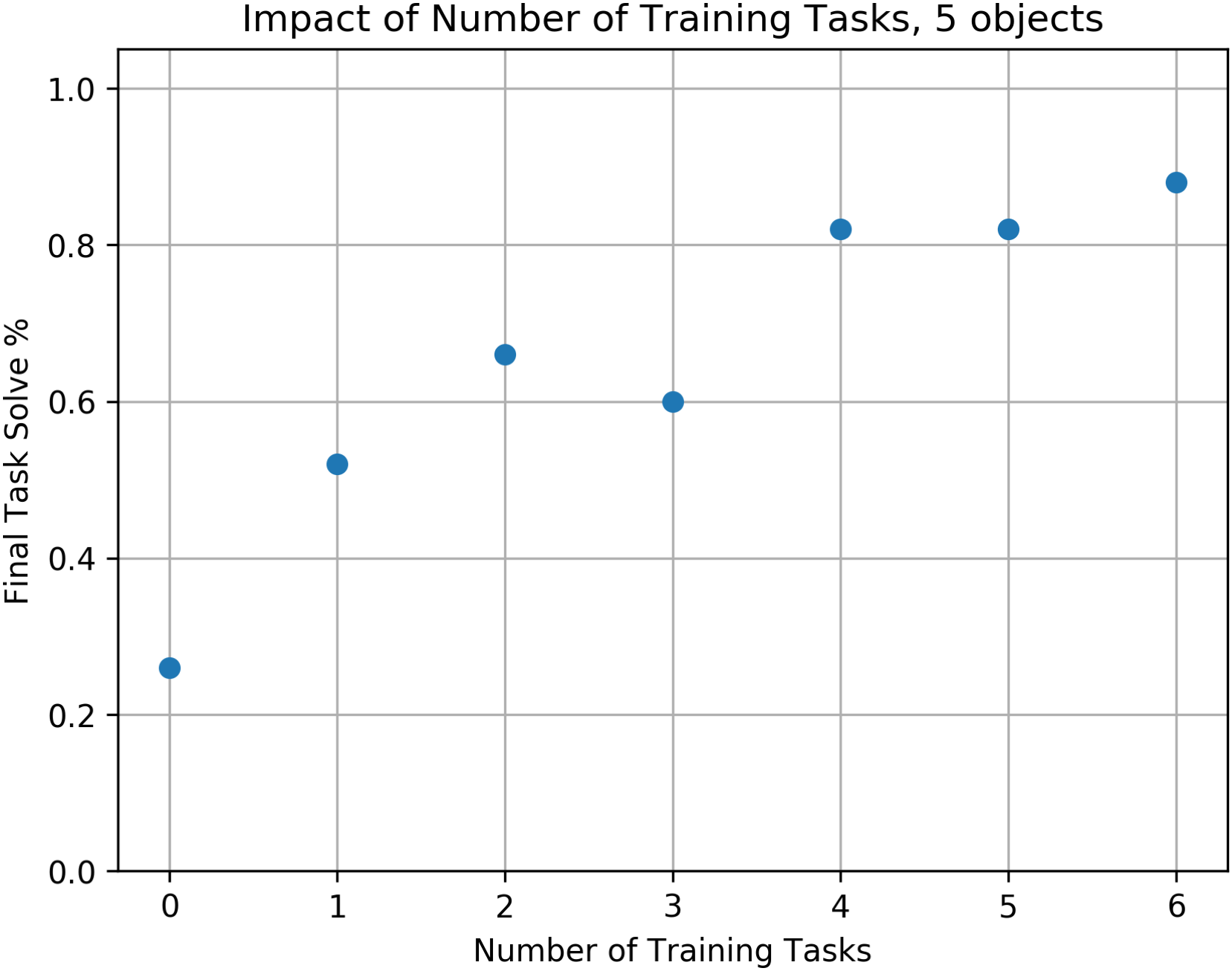}
    \caption{\emph{Left:} Final task learning curves (smoothed),
      showing that learning to do the evaluation tasks is much more
      efficient when starting from meta-trained specializers (blue,
      orange) versus a randomly initialized \ad\ learner
      (brown). \emph{Right:} To investigate the importance of having a
      diversity of training tasks, we ran the \ad\ learner while
      withholding some training tasks out of our full suite of 6. We
      can see that the final task performance (across all 3 evaluation
      tasks) improves when the agent is trained across more tasks.}
  \label{fig:finaltasklearningcurves}
\end{figure}

\figref{fig:finaltasklearningcurves} (left) shows the benefit of
learning in the final task when starting from meta-trained
specializers. The specializers meta-learned using the \ad\ learner
start off slightly worse than those meta-learned using the \subs\
learner, likely because the search space is larger (recall that the
\ad\ learner uses all the specializers), so timeouts are more
frequent. After some adaptation on the final task, the \ad\ learner
performs better. \figref{fig:finaltasklearningcurves} (right) suggests
that when the agent has access to more training tasks, it meta-learns
specializers that lead to better final task performance, given a fixed
amount of data in this final task. This is likely because each new
training task allows the agent to learn more about how to adapt its
specializers to the various object geometries.

\section{Conclusion and Future Work}
\label{sec:conclusion}

We used modular meta-learning to address the problem of learning
continuous action parameters in multi-task \tamp{}.

One interesting avenue for future work is to allow the specializers to
be functions of the full plan skeleton, which would provide them with
context necessary for picking good parameter values in more complex
domains. Another is to remove the assumption of deterministic
specializers by having them either be stochastic neural networks or
output a distribution over the next state, reparameterized using
Gumbel-Softmax~\cite{gumbelsoftmax}. Finally, we hope to explore tasks
requiring planning under uncertainty. These tasks would require more
sophisticated compositional structures; we would need to search over
tree-structured policies, rather than sequential plans as in this
work. This search could be made tractable using heuristics for solving
\pomdp s~\cite{sarsop,pbvi,despot}.



\section*{Acknowledgments}
We gratefully acknowledge support from NSF grants 1420316, 1523767,
and 1723381; from AFOSR grant FA9550-17-1-0165; from Honda Research;
and from Draper Laboratory. Rohan is supported by an NSF Graduate
Research Fellowship. Any opinions, findings, and conclusions expressed
in this material are those of the authors and do not necessarily
reflect the views of our sponsors.

\bibliography{references}

\end{document}